\documentclass[num-refs,dvipsnames,left=10cm]{wiley-article} %
\usepackage[disable]{todonotes}

\usepackage{siunitx}

\let\oldalign\align
\let\oldendalign\endalign
\renewenvironment{align}
  {\linenomathNonumbers\oldalign}
  {\oldendalign\endlinenomath}

\usepackage{ulem} %
\usepackage{tcolorbox} %
\usepackage{hyperref}

\usepackage{tagging} %

\hypersetup{
    colorlinks=true,
    linkcolor=blue,
    filecolor=magenta,      
    urlcolor=cyan,
    citecolor=gray,
    }

\usepackage{soulutf8}

\usepackage{amsmath} %

\usepackage{enumitem}

\usepackage{subcaption}

\usepackage{pdfpages} %

\usepackage{booktabs} %

\usepackage{caption}

\usepackage{xparse}

\usepackage{soulutf8}

\usepackage{stfloats}

\usepackage{floatrow}
\newfloatcommand{capbtabbox}{table}[][\FBwidth]

\usepackage{xargs}                      %

\papertype{Original Article}
\paperfield{Journal Section}

\title{Auto-Encoder Neural Network Incorporating X-Ray Fluorescence Fundamental Parameters with Machine Learning}

\author[1]{Matthew Dirks}
\author[1]{David Poole}

\affil[1]{Computer Science, University of British Columbia, Vancouver, British Columbia, V6T 1Z4, Canada}

\corraddress{Matthew Dirks}%
\corremail{mcdirks@cs.ubc.ca}

\fundinginfo{NSERC Discovery Grant;

MITACS Accelerate Internship: IT04336}

\runningauthor{Matthew Dirks et al.}

\begin{document}

\begin{tikzpicture}[remember picture,overlay]
	\node[anchor=south,yshift=10pt,draw] 
	at (current page.south) 
	{\parbox{\dimexpr\textwidth-\fboxsep-\fboxrule\relax}{
	\textit{Preprint 2023-01-19. Published version at \url{https://doi.org/10.1002/xrs.3340}}
	}};
\end{tikzpicture}%

\newcommand{\myinfobox}[1]{
    \marginpar{
        \tcbox[colframe=green!50!white,colback=white,width=100pt,grow to left by=-14.5cm]{ %
            \begin{minipage}{70pt}#1\end{minipage} %
        }
    }
}

\newcommand{\new}[1]{{\color{Green}#1}}
\newcommand{\reword}[1]{{\color{RubineRed}#1}}

\newcommand{\delete}[1]{{\color{WildStrawberry}--MARKED FOR REMOVAL\{#1\}--}} %

\newcommand{\grey}[1]{{\color{Gray}#1}}
\newcommand{\green}[1]{{\color{darkergreen}#1}}
\newcommand{\blue}[1]{{\color{blue}#1}}
\newcommand{\red}[1]{{\color{red}#1}}

\iftagged{markrevisions,markMD}{
    \newcommand{\mytest}[1]{MYTEST SAYS YES ##1}
    \newcommand{\revision}[1]{{\color{red}\uline{##1}}} %
    \newcommand{\strikeout}[1]{{\color{red}\sout{##1}}} %

    \iftagged{markMD}{ %
        \newcommand{\why}[2]{{\color{Plum}####1}\todof{WHY: ####2}}

        \newcommand{\mytodo}[1]{{\color{ForestGreen} --TODO\{####1\}--}}
    }{ %
        \newcommand{\why}[2]{####1}
        \newcommand{\mytodo}[1]{}
    }
}{ %
    \newcommand{\mytest}[1]{MYTEST SAYS NO ##1}
    \newcommand{\revision}[1]{{##1}} %
    \newcommand{\strikeout}[1]{{}} %
    \newcommand{\why}[2]{##1}
    \newcommand{\mytodo}[1]{}
}

\newcommand{\best}{\cellcolor{blue!25}} 

\newcommand{\hlgreen}[1]{{\hl{#1}}}
\DeclareRobustCommand{\hlcyan}[1]{{\sethlcolor{cyan}\hl{#1}}}

\newcommand{\subscript}[2]{$#1 _ #2$}

\renewcommand{\L}[1]{{L$_#1$}}

\newcommand{\knowledge}[1]{\subscript{K}{{#1}}} 

\newcommand{\numSimElements}[0]{51}
\newcommand{\numTargetElements}[0]{48}
\newcommand{\numEncodingVars}[0]{52} %

\newcommand{\Kalpha}[0]{K-\subscript{L}{3} (K$\alpha$)}
\newcommand{\mLR}[0]{\textbf{LR}}
\newcommand{\mLASSO}[0]{\textbf{LASSO}}
\newcommand{\mFCNN}[0]{\textbf{FCNN}}
\newcommand{\mCNN}[0]{\textbf{CNN}}
\newcommand{\mAXS}[0]{\textbf{AXS}}

\newcommand{\modEncoder}[0]{\textbf{Encoder}}
\newcommand{\modXRFSimulator}[0]{\textbf{XRF Simulator}}
\newcommand{\modOutput}[0]{\textbf{Output}}

\newcommandx{\mynote}[1]{\todo[backgroundcolor=cyan!25,bordercolor=cyan,noline]{#1}}
\newcommandx{\todof}[1]{\todo[fancyline,linecolor=red,backgroundcolor=red!25,bordercolor=red,]{#1}}
\newcommandx{\todox}[2][1=]{\todo[fancyline,linecolor=red,backgroundcolor=red!25,bordercolor=red,#1]{#2}}
\newcommandx{\unsure}[2][1=]{\todo[linecolor=red,backgroundcolor=red!25,bordercolor=red,#1]{#2}}
\newcommandx{\change}[2][1=]{\todo[linecolor=blue,backgroundcolor=blue!25,bordercolor=blue,#1]{#2}}
\newcommandx{\info}[2][1=]{\todo[linecolor=OliveGreen,backgroundcolor=OliveGreen!25,bordercolor=OliveGreen,#1]{#2}}
\newcommandx{\improvement}[2][1=]{\todo[linecolor=Plum,backgroundcolor=Plum!25,bordercolor=Plum,#1]{#2}}
\newcommandx{\thiswillnotshow}[2][1=]{\todo[disable,#1]{#2}}

\newcommand\todoin[2][]{\todo[inline, caption={2do}, #1]{\begin{minipage}{\textwidth-4pt}#2\end{minipage}}}

\DeclareRobustCommand{\todoL}[2]{{~\\\colorbox{yellow}{\begin{minipage}{\linewidth}#2\end{minipage}}\todo[fancyline,linecolor=cyan,backgroundcolor=cyan!25,bordercolor=cyan,]{#1}}}

\begin{frontmatter}
\maketitle

\begin{abstract}

We consider energy-dispersive X-ray Fluorescence (EDXRF) applications where the fundamental parameters method is impractical
such as when instrument parameters are unavailable.
For example, on a mining shovel or conveyor belt, 
rocks are constantly moving (leading to varying angles of incidence and distances)
and there may be other factors not accounted for (like dust). 
Neural networks do not require instrument and fundamental parameters
but training neural networks requires XRF spectra labelled with elemental composition, which is often limited because of its expense.
We develop a neural network model that 
learns from limited labelled data
\revision{and also benefits from domain knowledge by learning}
to invert a forward model. 
The forward model uses transition energies and probabilities of all elements
and parameterized distributions to approximate other fundamental and instrument parameters.
We evaluate the model and baseline models on a rock dataset from a lithium mineral exploration project.
\revision{Our model works particularly well for some low-Z elements (Li, Mg, Al, and K)
as well as some high-Z elements (Sn and Pb) despite these elements being outside the suitable range for common spectrometers to directly measure, likely owing to the ability of neural networks to learn correlations and non-linear relationships.}

\end{abstract}%

\end{frontmatter}

\section{Introduction}

The fundamental parameters method is often used to quantify elemental composition because
the underlying physical principles are well-understood in X-ray fluorescence (XRF) 
\cite{deboer1993howAccurateIsFP,lachance1995xrfBook}.
We consider XRF applications where the fundamental parameters method is impractical,
typically because instrument parameters are unavailable or they vary from instance to instance.
For example, in sensors out in the field, such as on a mining shovel or conveyor belt \cite{bamberDirks2016shovel}, 
rocks are constantly moving (leading to varying angles of incidence and distances).
There may also be other factors not accounted for (such as dust) which are difficult to model.
Neural networks have been used as an alternative \cite{Kaniu2011chemometricEDXRF,li2019AnnXrf,Jones2022pigmentsSynthetic} to the fundamental parameters method.
However, neural networks require labelled data but obtaining spectra with corresponding elemental composition is often expensive.
For example, in mining, geochemical assay requires grinding and crushing large rock samples which is an expensive process.

Aside from training with more data, performance of machine learning models can also be improved by
incorporating domain knowledge;
this is an attractive way to improve generalizability \cite{karpatne2017theoryGuided,gulccehre2016knowledgeMatters}. %
Utilization of fundamental parameters within a neural network has only been achieved indirectly.
For instance, early work \cite{luo2002algCombiningNNwithFP} calculates theoretical relative intensities of each element using fundamental parameters (and instrument parameters). 
A small fully-connected neural network (with 1 hidden layer of size 10 and 1 output) maps these intensities to concentrations. %
This network does not benefit from machine learning advancements made in the last 20 years.
Recent work \cite{Jones2022pigmentsSynthetic} trains a convolutional neural network 
on simulated spectra generated using fundamental parameters before
fine-tuning the model on real spectra.
Our approach differs from both of these in that we incorporate a forward model,
that simulates a spectrum given the elemental composition,
directly in a neural network.

  The proposed neural network learns the inverse of the fundamental parameters method
  alongside 
  simulator parameters used to approximate the effects of the instrument and environment. 
  Fundamental parameters, specifically transition energies and probabilities,
  are built into the simulator.
The specific neural network architecture we use is an auto-encoder where 
the encoder transforms spectra to a lower-dimensional encoding representing properties of the rocks
and the decoder translates properties of rocks back into spectra through a trainable simulator.
This style of auto-encoder is an implementation of analysis-by-synthesis \cite{halle1962speech}.
\why{}{strikeout text moved to methods section}%
The model is tested on a rock dataset from a lithium mineral exploration project to demonstrate its potential.

\revision{We found that the analysis-by-synthesis model outperforms the baselines and other neural networks on 11 elements including several low-Z elements (Li, Mg, Al, and K) and high-Z elements (Sn and Pb) despite these elements being outside the suitable range for common spectrometers to directly measure, likely owing to the ability of neural networks to learn correlations and non-linear relationships combined with domain knowledge in the forward model.}

\section{Dataset}

177 rock core samples (about 3 inches in length) obtained from a lithium mineral exploration project
were analyzed by x-ray fluorescence and then sent for geochemical assay.

\paragraph{XRF Spectra:}

Energy-dispersive X-ray fluorescence (EDXRF) is used to collect spectra from 
rock samples (which are not prepared or processed in any way).
Each rock sample is analyzed under 4 orientations
and the 4 spectra are averaged together.
\revision{The XRF spectrometer used in this study produces a spectrum with 1024 channels. 
It uses a 50 kV X-ray tube with a silver (Ag) target. The detector uses a beryllium (Be) window.
The spectrometer is automatically moved to be approximately 10 cm away from the rock sample using a laser distance sensor.}

\paragraph{Geochemical Analysis:}
Geochemical analysis is performed for each rock sample by an assay lab to determine the concentrations of \numTargetElements{} elements\footnote{\revision{Elements are Ag, Al, As, Ba, Be, Bi, Ca, Cd, Ce, Co, Cr, Cs, Cu, Fe, Ga, Ge, Hf, In, K, La, Li, Mg, Mn, Mo, Na, Nb, Ni, P, Pb, Rb, Re, S, Sb, Sc, Se, Sn, Sr, Ta, Te, Th, Ti, Tl, U, V, W, Y, Zn, and Zr.}}. 
The assays are destructive; the analysis uses 4-acid digestion followed by inductively coupled plasma mass spectrometry (ICP-MS). 
\revision{Analysis performed by \href{alsglobal.com}{ALS} using their proprietary ME-MS61 method \cite{ALS}.} 
This method provides gold-standard composition estimates.

\section{Models}

Prediction models, such as linear regression or neural networks, require data for training.
In this study, the training data set consists of a spectrum per rock labelled with corresponding composition provided by geochemical analysis. 
Labelled data is often limited because 
of the excessive cost and time, and destructive nature, of geochemical analysis.
Limited labelled data is challenging because highly-parameterized models, 
such as neural networks,
are prone to overfitting.
We test a range of models with different numbers of parameters and different regularization \cite{kuk2017regularizationTaxonomy} schemes, explained below.

\revision{All models are evaluated by 10-fold cross-validation \cite{fearn2008CV}.
That is, the dataset is randomly shuffled once then partitioned into 10 ``folds''.
For each fold, each model is tested on the samples in the fold
and trained on the remaining 9 folds.
The average prediction error (MSE) across the 10 folds is an estimate of generalization ability
with standard error (SE) of this prediction estimated as:
$\frac{1}{\sqrt{k}} SD\{CV_{1}, ..., CV_{k}\}$
where 
$k$ is 10,
$SD$ is the standard deviation,
and $CV_i$ is the mean squared error (MSE) of the $i\textsuperscript{th}$ fold.
}

\subsection{LR}
\textit{Simple linear regression}, which we abbreviate as \mLR{}, is a model with one input and one output variable.
For each element, we build a linear model from the photon count at the \Kalpha{} line for that element to that element's concentration.
This model has the fewest parameters of the models tested in this paper.

\subsection{LASSO}
Least Absolute Shrinkage and Selection Operator (\mLASSO{}) is a linear regression model
that optimizes squared error plus 
the sum of the absolute values of the model's coefficients (\L{1}-norm on the coefficients) \cite{tibshirani1996lasso}
times a constant.
The inputs are the full spectrum.
This model has been used in spectroscopy \cite{ojelund2001-NIR-LASSO,boucher2015libs} to induce sparsity on the large feature spaces exhibited by most spectra.
The \L{1} regularizer essentially selects which channels of the spectra should participate in the regression model.
We train a \mLASSO{} model for each element.

\subsection{FCNN}

Three neural networks are considered,
of which the first
is a single-layer Fully-Connected Neural Network (\mFCNN{}).
The model's input, a spectrum, is directly connected to \numTargetElements{} outputs representing element concentrations.
A RELU \cite{Mercioni2020activationFunctions} activation function is used on the output layer to clip negative values to zero.
As per standard practice, early-stopping, dropout, and \L{1} regularization are used to reduce over-fitting \why{\cite{hutter2021welltuned}}{Gives overview of reg techniques (and a method for optimizing them).}.

\revision{For all the neural networks tested in this paper, 
grid search found the following hyperparameters:
learning rate is 0.001, early-stopping patience is 1000 epochs, L$_1$ regularization factor is 0.001, and dropout starts at 50\% probability and gradually backs off until dropout is disabled at epoch 10000.}
Training is run until early-stopping criteria is met. %
\revision{FCNN trained for 16000 epochs on average.}
All the neural networks were programmed in Python version 3.6 using \href{https://tensorflow.org/}{TensorFlow} version 1.13.

\subsection{CNN}
The second neural network is a Convolutional Neural Network (\mCNN{}) \cite{lecun1998CNN}.
This type of model is common in other spectroscopy domains \cite{acquarelli2017cnnForVib,lui2017CNNRaman,malek20181dCNN,fearn2018modernCNN,fan2019raman,chatzidakis2019sciReport,zhang2019DeepSpectra,yang2019progressAndGuide,Mishra2021mangoes} but not in XRF.
Convolution layers incorporate some general domain knowledge about spectral data.
Specifically, a convolution layer captures the notion that features are likely to be locally correlated.
This is true for XRF spectra because neighbouring channels' intensities are correlated 
within the background continuum,
and within peaks due to limited detector resolution \cite{beckhoff2007handbookXRF}.

\begin{figure*}[tbh]
  \vspace*{0cm}
  \makebox[\linewidth]{
    \includegraphics[width=1\linewidth]{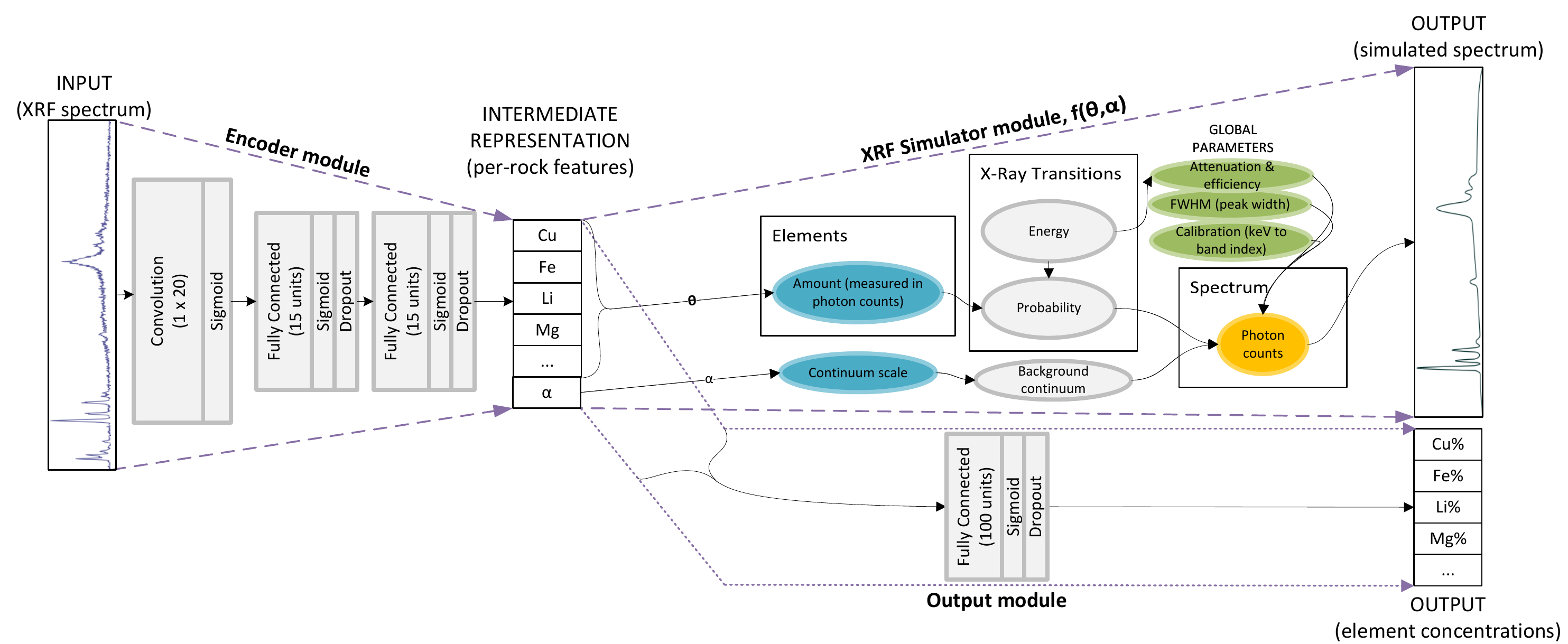}
  }
  \caption{Architecture of Analysis-by-XRF-Synthesis (\mAXS{}) model. The \mCNN{} architecture is equivalent to the \modEncoder{} and \modOutput{} modules.}
  \label{fig:xrf_ae}
\end{figure*}

The \mCNN{}'s architecture, shown in Figure \ref{fig:xrf_ae}, is equivalent to the \modEncoder{} module followed by the \modOutput{} module.
All the layers use sigmoid \cite{Mercioni2020activationFunctions} activation functions, including on the output, which allows the network to learn non-linearities.
The fully-connected layers learn the dependencies between features that may be present in the data.
Early-stopping, dropout, and \L{1} regularization are used to reduce over-fitting.
\revision{This model trained for 12000 epochs on average.}

\subsection{Analysis-by-Synthesis}

The final model is an implementation of analysis-by-synthesis, which we call Analysis-by-XRF-Synthesis (\mAXS{}),
using an auto-encoder neural network.
Auto-encoders \cite{Hinton2006AE,aggarwal2018textbookML} are commonly used to build low-dimensional informative representations \cite{le2012highLevelFeatures,bengio2013representation}.
\revision{%
Analysis-by-synthesis has been used in computer vision, for instance, where a graphics engine generates images and the neural network learns the inverse process which translates images into properties of physical objects \cite{tieleman2014thesis,wu2017visualDeAnimation}.
}%
\mAXS{} consists of 3 modules (shown in Figure \ref{fig:xrf_ae}): \modEncoder{}, \modXRFSimulator{}, and \modOutput{}.
The decoder of a typical auto-encoder is replaced by the \modXRFSimulator{} module (see below).

\subsubsection{XRF Simulator Module}
\label{app:xrfsim}

The \modXRFSimulator{} module (shown in Figure \ref{fig:xrf_ae}) is a forward model using fundamental parameters.
Using the TensorFlow software library, the \modXRFSimulator{} is programmed to be fully-differentiable;
differentiation is a requirement of the backpropagation algorithm used to train neural networks.

The paths of individual photons are not simulated, 
but rather the expected histogram of photon counts for each energy (bin) using known X-ray transition energies and associated probabilities.
X-ray transition energies were downloaded from NIST's X-Ray Transition Database\footnote{Directly measured experimental transition energies downloaded from the National Institute of Standards and Technology (NIST) X-Ray Transition Database (\url{https://www.nist.gov/pml/x-ray-transition-energies-database}) \href{http://physics.nist.gov/PhysRefData/XrayTrans/Html/search.html}{here}.} 
and the probabilities of each transition are from the Evaluated Atomic Data Library\footnote{The 1997 release of the Evaluated Atomic Data Library (EADL97) is available from Nuclear Data Services of the International Atomic Energy Agency at \url{https://www-nds.iaea.org/epdl97/libsall.htm}.}.
All K, L1, L2, and L3 transition types are included in the model.
We refer to the set of transitions for element $e$ as $T_e$ and the energy and probability for transition $t$ as $t_{i}$ and $t_{p}$ respectively.

Firstly, a spectrum is modelled as a function\footnote{%
The notation ``$\text{name}\colon a, b, c \mapsto x$'' denotes a function named ``name'' with 3 arguments (a, b, c) that returns $x$.%
} from energy, $i$, to photon count (or intensity):
\begin{equation}
  \text{spectrum}\colon i \mapsto \text{photon count}
\end{equation}
The peak caused by a transition is modelled as a Lorentzian function%
\footnote{\revision{%
  A Lorentzian has a similar shape to a Gaussian but is more narrow around the peak with longer tails.
  Another profile may also fit well, such a Gaussian or Voigt function. Regardless, we only need an approximate fit because, ultimately, the \modEncoder{} will learn whatever form best explains the data.%
}}%
, $L$, which outputs a spectrum:
\begin{align}
  L&\colon t_{i}, t_{p}, \Gamma \mapsto \text{spectrum} \\
  L(t_{i}, t_{p}, \Gamma)(i) &= \frac
    {t_{p}  \left(\frac{\Gamma}{2}\right)^2}
    {(i - t_{i})^2 + \left(\frac{\Gamma}{2}\right)^2}
\end{align}
where
$t_{i}$ is the location of the center of the Lorentzian peak (given by the energy of transition $t$),
$t_{p}$ is the height of the peak (given by the probability of transition $t$),
and
$\Gamma$ is the width of the peak.

A predicted spectrum, $g$, for some sample with a given composition, $\theta$, is generated by summing up
the spectra, channel-wise, produced by all the transitions of all the elements:
\begin{align}
  \boldsymbol{\theta} &= (\theta_\text{Cu}, \theta_\text{Fe}, \text{etc}) \\
  g&\colon \boldsymbol{\theta} \mapsto \text{spectrum} \\
  g(\boldsymbol{\theta})(i) &= \sum_{e\in{E}}\Big(\theta_e \sum_{t\in{T_e}} L(t_{i}, t_{p}, \Gamma)(i) \Big)
\end{align}
where $\theta_e$ is the amount of scaling needed to scale $L$ into photon counts.

The \modXRFSimulator{} obviates the need to specify instrument and environment properties by approximating instrument-specific and environment-specific effects (such as background continuum, attenuation, and efficiency).
First, 
attenuation and efficiency curves
are jointly represented by the product of two sigmoid curves, approximated by
\begin{align}
  S&\colon a_1, c_1, a_2, c_2 \mapsto \text{spectrum} \\
  S(a_1, c_1, a_2, c_2)(i) &= \frac{1}{1 + 
    e^{a_2(c_2 - i)} + 
    e^{a_1(c_1 - i)}}
\end{align}
$S$ is used to automatically calibrate for the specific instrument based on the data
and is parameterized by four global parameters ($a_1$, $c_1$, $a_2$, $c_2$) that are learned.
The last piece is the background continuum which is added to the final spectrum.
Background continuum is approximated 
as a B\'ezier curve, $b$, 
which is a function of two global parameters, $p_1$ and $p_2$, and $\alpha$ that depends on the rock sample:
\begin{align}
  b&\colon p_1, p_2, \alpha \mapsto \text{spectrum} \\
  b(p_1,p_2,\alpha)(i) &= \alpha \big(
    3 p_1 i (1-i)^2 +
    3 p_2 i^2 (1-i)
    \big)
\end{align}
Finally, the complete spectrum for a sample is produced from its composition, $\theta$, and amount of background, $\alpha$,
by generating a theoretical spectrum, $g$, scaling it by $S$, and adding the background continuum, $b$:
\begin{align}
  f&\colon \theta \mapsto \text{spectrum} \\
  f(\theta, \alpha)(i) &= 
      g(\theta)(i) \times S(a_1,c_2,a_2,c_2)(i)
    + b(p_1,p_2,\alpha)(i)
  \label{eq:sim}
\end{align}
This function, $f$, constitutes the \modXRFSimulator{} module and is depicted graphically in Figure \ref{fig:xrf_ae}.
Note that $t_{i}$ and $t_{p}$ are known from fundamental parameters and $\Gamma$, $a_1$, $c_1$, $a_2$, $c_2$, $p_1$, $p_2$, and $\alpha$ are learned to fit data.

\subsubsection{Output Module}
  The intermediate representation ($\theta$, $\alpha$) is used as input to another learning module, the \modOutput{} module (shown in Figure \ref{fig:xrf_ae}).
  It consists of a fully-connected layer with \L{1} regularization, a sigmoid activation function, and dropout.
  The output is the proportion of each element (unlike $\theta$ in the intermediate representation which contains multipliers for each element).
  All \numTargetElements{} elements are predicted simultaneously, in the one model.
  \revision{AXS trained for 55000 epochs on average.}

\subsubsection{Objective Function}

    As in an auto-encoder, a reconstruction error, $\mathcal{J}_{r}$, is minimized causing the \modEncoder{} and \modXRFSimulator{} modules to learn to reconstruct the training spectra.
    $\mathcal{J}_{r}(x^{}_s,x^\prime_s)$ is the mean squared error (MSE) between observed, $x^{}_s$, 
    and reconstructed, $x^\prime_s$, spectra:
\begin{equation}
  \mathcal{J}_{r}(x^{}_s,x^\prime_s) = \frac{\sum_{i=1}^{m}\big(x^{}_{si} - x^\prime_{si}\big)^2}{m}
\end{equation}
where 
$x^{}_{si}$ and $x^\prime_{si}$ are the $i^{th}$ channel of the observed and reconstructed spectra of sample $s$ respectively,
and $m=1024$ is the number of channels in the spectrum (indexed by $i$).
$x^\prime_s$ is the reconstructed spectra produced by the \modXRFSimulator{} function, $f$.

Prediction error, $\mathcal{J}_{p}$, penalizes the difference between predicted concentrations and ground truth (given by geochemical analysis):
\begin{equation}
  \mathcal{J}_{p}(y^{}_s,y^\prime_s) = \frac{\sum_{e\in{E}}  \big(y^{}_{se} - y^\prime_{se}\big)^2}{|E|}
\end{equation}
where
$y^{}_{se}$ and $y^\prime_{se}$ are the actual and predicted (respectively) concentrations of element $e$ and sample $s$,
$E$ is the set of all elements in use,
and $|E|$, the number of elements, is \numTargetElements{}.
$y^\prime_s$ is produced by the \modOutput{} module.
Finally, a weighted sum of the reconstruction error and prediction error yields the loss function, $\mathcal{J}(w)$, used to train the model:
\begin{equation}
  \mathcal{J}(w) = \sum_{s} \mathcal{J}_{p}(y^{}_s,y^\prime_s) + \beta \mathcal{J}_{r}(x^{}_s, x^\prime_s)
\end{equation}
where 
$s$ is a sample,
$w$ is the set of all tunable weights (which are the neural network weights from the \modEncoder{} and 
\modOutput{} modules and the global parameters from the \modXRFSimulator{})
and
$\beta$ is the weight of the reconstruction loss (which is a hyperparameter).
Optimizing $\mathcal{J}_{r}$ and $\mathcal{J}_{p}$ together has been shown to be better than or equal to optimizing prediction error, $\mathcal{J}_{p}$, alone \cite{Le2018SupervisedAE}.

\section{Results}

\numTargetElements{} elements were included in geochemical analysis, 30 of these were removed from the results because
none of the methods were able to do better than predicting the mean; \revision{these 30 elements have very low-abundance and little variation in the samples studied, making the results inconclusive for these elements}. 
Of the remaining 18,
\mAXS{} achieved the best MSE on 11 elements, summarized in Table \ref{tbl:eval_count}. 
A full list of expected prediction errors and standard errors for all 18 elements are given in Appendix \ref{app:fullEval}. 
Note that \mLR{} is not applicable to low-Z elements, so no results are reported for these;
low-Z elements are particularly challenging for several reasons including small excitation factors, high attenuation, and increased scattering (\cite{lachance1995xrfBook}, page 204).
\revision{Li is of particular interest in this project;}
\mAXS{} achieved the best prediction error on lithium (Li, Z=3) as shown in Figure \ref{fig:errorbars_Li}.
\revision{From the results (Appendix \ref{app:fullEval} Figure \ref{fig:errorbars}) we see that \mAXS{} outperformed the baselines and other neural networks on several low-Z elements (Li, Mg, Al, and K) and high-Z elements (Sn and Pb) despite these elements being outside the suitable range for the spectrometer to directly measure.}
\revision{Gallium (Ga) is well within range for the spectrometer but did poorly when calibrating against the \Kalpha{} peak directly (using \mLR{}), whereas the multivariate models (\mLASSO{}, \mFCNN{}, and \mCNN{}) do much better and \mAXS{} does the best.
For the remaining elements (where \mAXS{} was not the best model), the results were roughly tied between the competing models.
}

We might expect \mLR{}, \mFCNN{}, \mCNN{}, and \mAXS{} to improve upon each other.
\mFCNN{} is expected to outperform \mLR{} because \mFCNN{} has more parameters and utilizes the whole spectrum (albeit, at the risk of overfitting);
\mCNN{} is expected to outperform \mFCNN{} because of the regularization power of the convolution layer;
and \mAXS{} is expected to outperform \mCNN{} because it incorporates domain knowledge.
Such monotonic improvement is observed in 
3 elements (Al, Ga, and Pb),
similar to the one shown in Figure \ref{fig:monotonic_improvement} (graphs for all elements are shown in Appendix Figure \ref{fig:errorbars}).
6 elements (Ca, Cs, Mo, S, Sb, and Sr) performed oppositely%
---with \mFCNN{}'s MSE less than all other neural networks---but standard error is large, suggesting performance could have gone either way. 
\mCNN{} did not have the best MSE on any elements, but it was the runner-up for 4 elements (Al, Ga, Pb, Sn) as can be seen in Appendix \ref{app:fullEval}.

\begin{figure*}[htb]\RawFloats\CenterFloatBoxes
\begin{floatrow}

  \ffigbox[\Xhsize/3]{
    \includegraphics[width=0.7\linewidth]{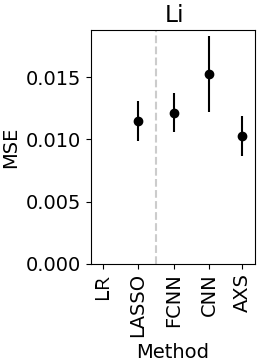}
    \caption{Example where \mAXS{} does well but \mCNN{} does not. 
    \revision{Results for \mLR{} are absent because Li does not have a \Kalpha{} line.}
    }
    \label{fig:errorbars_Li}
  }

  \ffigbox[\Xhsize/2]{
    \includegraphics[width=0.7\linewidth]{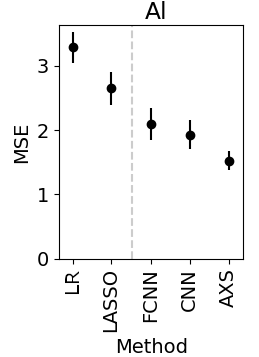}
    \caption{Example of monotonic improvement in MSE. Error bars are $\pm$ 1 Standard Error.}
    \label{fig:monotonic_improvement}
  }

  \definecolor{mylightgrey}{gray}{0.95}
  \ttabbox{
    \rowcolors{2}{gray!10}{white}
    \resizebox{0.43\Xhsize}{!}{%
    \begin{tabular}{ll}
      \toprule
      Model & Best on \\
      \midrule
                    \mLR{} & S \\
                 \mLASSO{} & Rb Sb \\
                  \mFCNN{} & Ca Cs Mo Sr\\
                   \mCNN{} &  \\
                   \mAXS{} & Al Be Fe Ga Hf K \\ 
      \rowcolor{white} & Li Mg Pb Sn Zr\\
      \bottomrule
    \end{tabular}}
    \caption{Elements where each model achieved the best MSE.}
    \label{tbl:eval_count}
  }

\end{floatrow}
\end{figure*}

\section{Discussion}

\begin{figure*}[tbp]
 \begin{center}
  \includegraphics[width=1\linewidth]{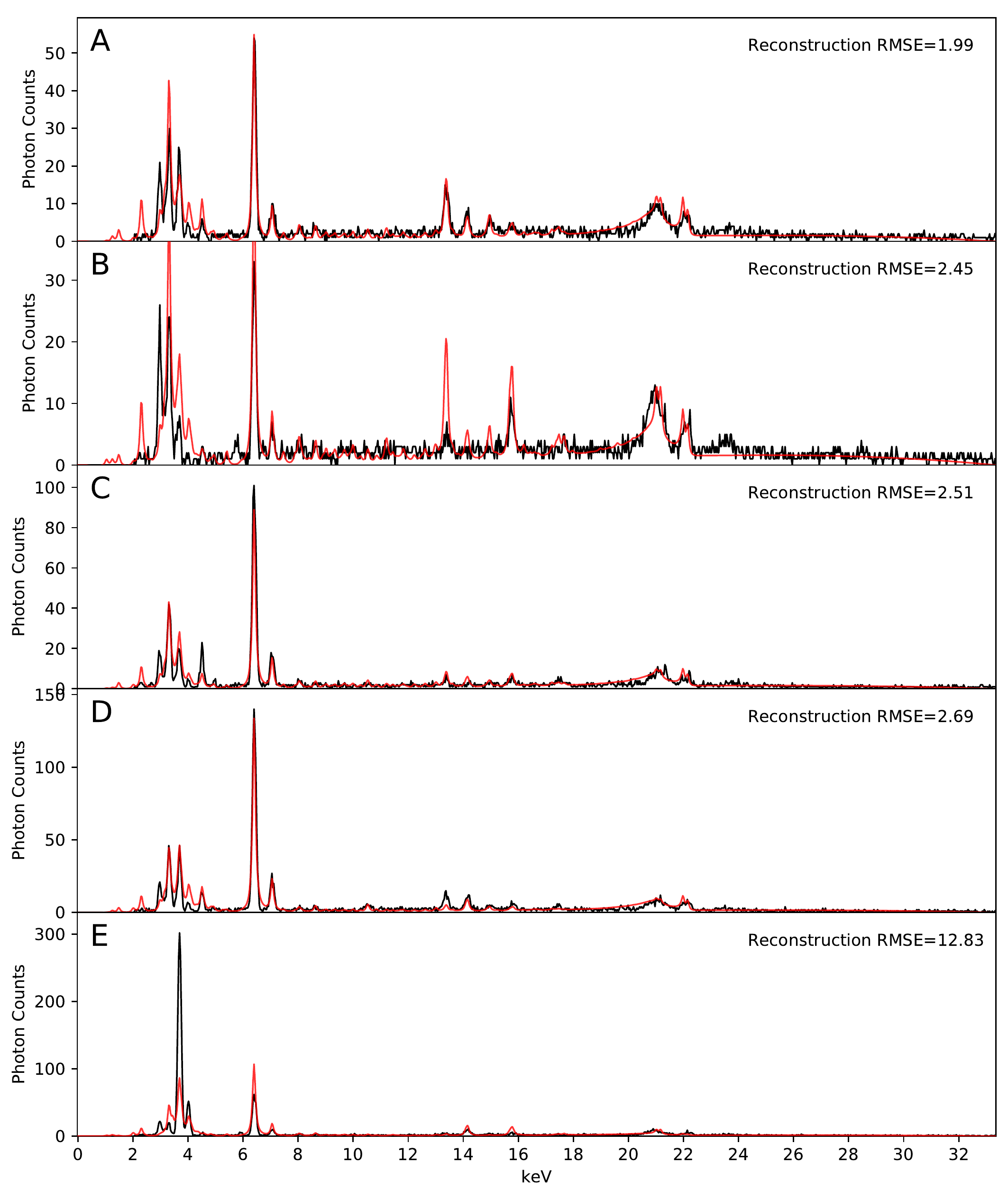}
  \caption{\revision{Example spectra and \mAXS{} reconstructions for 5 different samples.
  \textbf{A} has the lowest RMSE and \textbf{E} has the highest RMSE.
  \textbf{B}, \textbf{C}, and \textbf{D} have RMSE near the median (which is 2.58).
  The mean RMSE is 3.01.
  }}

  \label{fig:reconstruction}
 \end{center}
\end{figure*}

From looking at spectral reconstructions we can gain an insight into what the model does well.
\revision{Example reconstructions} from \mAXS{}'s auto-encoder is shown in Figure \ref{fig:reconstruction} (in red).
\revision{Reconstruction RMSE for the majority of samples is around the median (2.58),
such as examples B, C, and D in Figure \ref{fig:reconstruction};} many of the peaks fit very well.
Peaks below 5 keV rarely fit well, \revision{such as in examples A, B, and E}.
This may be due to interference and increased photon scattering from the rough rock surface. 
Improving the simulator to account for secondary fluorescence (as in a recently published XRF simulator \cite{brigidi2017GIMPyXRFsim}), sum peaks, diffraction peaks, escape peaks, and other artificial peaks \cite{tanaka2017artificialPeaksXRF} may help the poor-fitting regions of the spectra and improve the reconstructions,
which may then lead to improved predictions too.
Implementing this functionality in a differentiable manner, as required for the method described in this paper, 
would be a good way to improve upon \mAXS{}.

\mAXS{} has the benefit of not requiring instrument parameters because it learns to model the XRF spectra from data. 
Since it relies on data,
\mAXS{} should be retrained if the instrument changes or if the distribution of distances, rock sizes, or mineralogy changes.
Another benefit of \mAXS{} is that inference is fast, requiring only a single pass through the \modEncoder{} module 
to make a prediction and it does so for all \numTargetElements{} simultaneously.
This is fast compared to peak-fitting routines that run an iterative optimization routine to fit each new observed spectrum.
Lastly,
a benefit of auto-encoder architectures, such as \mAXS{}, is the ability to train in a semi-supervised \cite{kaneko2018semisuper} fashion;
the \modEncoder{} and \modXRFSimulator{} can be trained on unlabelled spectra 
while the whole model is trained on spectra labelled with composition provided by geochemical analysis.
For example,
in a mining shovel or conveyor belt application, it is expected that few samples will be sent for geochemical analysis 
(because large samples are more expensive)
but spectra alone will be relatively cheap to obtain. 
\revision{Therefore, much of \mAXS{} can be trained on unlabelled data thus reducing the need to obtain many geochemical analyses.}

Other variants of neural network architectures may also improve performance, but remain to be investigated, such as 
using deconvolutional layers (in the decoder) \cite{aggarwal2018textbookML}, %
variational auto-encoders \cite{kingma2019introToVAE},
and others.
The analysis-by-synthesis method explored in this paper may also be applicable to other spectroscopy disciplines,
assuming the XRF spectra simulator is replaced with an appropriate simulator.

\section{Conclusion}

The problem of XRF spectroscopy quantification where labelled data is limited 
and the fundamental parameters method is not directly applicable was investigated. 
As a proof-of-concept, an analysis-by-synthesis style auto-encoder trained on rock core samples demonstrated the potential of this method.
We combined (1) learning from limited labelled data with a neural network and (2) an XRF spectra simulator based on fundamental parameters.
In experimental results, we observed improved predictions on 11 elements, 
\revision{6 of which fall outside the ideal range of the XRF spectrometer}. 
We are confident that this method can be further refined and will extend the reach of XRF to more difficult applications.

\section*{acknowledgements}
We thank MineSense Technologies (\url{https://minesense.com/}) and Mitacs (\url{https://www.mitacs.ca/}) for supporting this research.

\printendnotes

\bibliography{references}

\appendix
\section{Appendix}

\subsection{Full Evaluation Results}
\label{app:fullEval}

\numTargetElements{} elements have ground truth, 30 of these were removed from the results because
none of the models were able to do better than predicting the average, likely due to low-abundance of these elements.
The results from the remaining elements are shown in Table \ref{tbl:eval} and Figure \ref{fig:errorbars}.
The removed elements are Ag, As, Ba, Bi, Cd, Ce, Co, Cr, Cu, Ge, In, La, Mn, Na, Nb, Ni, P,  Re, Sc, Se, Ta, Te, Th, Ti, Tl, U,  V,  W,  Y,  and Zn.

\begin{table}[htp]
  \tiny
  \centering
    \rowcolors{2}{white}{gray!10}
    \texttt{%
      \begin{tabular}{l|ll|llll}
\toprule
       & \multicolumn{2}{c|}{Baselines} & \multicolumn{3}{c}{Neural Networks} \\
\midrule
\rowcolor{white}
{}  &                                             LR &                                    LASSO &                                     FCNN &                                      CNN  &                                      AXS \\
Element   &                                          &                                          &                                          &                                           &                                          \\
\midrule
  Al      &       3.29e+0{\color{gray}$\pm$2.4e-1}   &       2.65e+0{\color{gray}$\pm$2.6e-1} &       2.09e+0{\color{gray}$\pm$2.5e-1} &       1.93e+0{\color{gray}$\pm$2.3e-1} &  \best1.52e+0{\color{gray}$\pm$1.5e-1} \\
  Be      &                                          &       1.25e-7{\color{gray}$\pm$2.2e-8} &       1.36e-7{\color{gray}$\pm$2.6e-8} &       1.33e-7{\color{gray}$\pm$2.9e-8} &       \best1.19e-7{\color{gray}$\pm$1.6e-8} \\
  Ca      &       8.96e+0{\color{gray}$\pm$2.5e+0}   &       1.11e+1{\color{gray}$\pm$2.6e+0} &  \best6.71e+0{\color{gray}$\pm$2.4e+0} &       1.44e+1{\color{gray}$\pm$5.1e+0} &       1.01e+1{\color{gray}$\pm$3.1e+0} \\
  Cs      &       1.93e-4{\color{gray}$\pm$4.5e-5}   &       1.42e-4{\color{gray}$\pm$3.0e-5} &  \best1.24e-4{\color{gray}$\pm$3.1e-5} &       1.43e-4{\color{gray}$\pm$3.3e-5} &       1.34e-4{\color{gray}$\pm$2.7e-5} \\
  Fe      &       7.85e-1{\color{gray}$\pm$1.6e-1}   &       1.01e+0{\color{gray}$\pm$1.3e-1} &       7.67e-1{\color{gray}$\pm$1.7e-1} &       8.33e-1{\color{gray}$\pm$1.6e-1} &       \best7.21e-1{\color{gray}$\pm$1.4e-1} \\
  Ga      &       3.19e-7{\color{gray}$\pm$2.2e-8}   &       2.31e-7{\color{gray}$\pm$1.8e-8} &       2.45e-7{\color{gray}$\pm$2.6e-8} &       1.95e-7{\color{gray}$\pm$2.0e-8} &  \best1.72e-7{\color{gray}$\pm$2.0e-8} \\
  Hf      &                                          &       1.56e-7{\color{gray}$\pm$2.1e-8} &       1.52e-7{\color{gray}$\pm$1.8e-8} &       2.12e-7{\color{gray}$\pm$4.6e-8} &  \best1.46e-7{\color{gray}$\pm$1.6e-8} \\
  K       &       1.10e+0{\color{gray}$\pm$1.5e-1}   &       1.28e+0{\color{gray}$\pm$1.2e-1} &       1.13e+0{\color{gray}$\pm$9.4e-2} &       1.29e+0{\color{gray}$\pm$2.8e-1} &  \best8.70e-1{\color{gray}$\pm$1.5e-1} \\
  Li      &                                          &       1.15e-2{\color{gray}$\pm$1.6e-3} &       1.22e-2{\color{gray}$\pm$1.6e-3} &       1.53e-2{\color{gray}$\pm$3.0e-3} &       \best1.03e-2{\color{gray}$\pm$1.6e-3} \\
  Mg      &       1.54e+1{\color{gray}$\pm$1.2e+0}   &       7.23e+0{\color{gray}$\pm$5.7e-1} &       5.73e+0{\color{gray}$\pm$8.0e-1} &       6.85e+0{\color{gray}$\pm$1.9e+0} &  \best4.42e+0{\color{gray}$\pm$5.4e-1} \\
  Mo      &       2.74e-5{\color{gray}$\pm$5.7e-6}   &       2.55e-5{\color{gray}$\pm$6.1e-6} &  \best2.41e-5{\color{gray}$\pm$6.0e-6} &       6.08e-5{\color{gray}$\pm$1.5e-5} &       5.92e-5{\color{gray}$\pm$1.4e-5} \\
  Pb      &                                          &       3.47e-7{\color{gray}$\pm$6.4e-8} &       3.22e-7{\color{gray}$\pm$5.3e-8} &       2.89e-7{\color{gray}$\pm$4.6e-8} &  \best2.63e-7{\color{gray}$\pm$4.1e-8} \\
  Rb      &       1.53e-4{\color{gray}$\pm$1.9e-5}   &  \best1.17e-4{\color{gray}$\pm$1.6e-5} &       1.32e-4{\color{gray}$\pm$1.8e-5} &       2.22e-4{\color{gray}$\pm$4.2e-5} &       1.18e-4{\color{gray}$\pm$1.5e-5} \\
  S       &  \best8.46e-1{\color{gray}$\pm$1.8e-1}   &       9.27e-1{\color{gray}$\pm$1.4e-1} &       9.30e-1{\color{gray}$\pm$2.0e-1} &       1.27e+0{\color{gray}$\pm$2.0e-1} &       1.20e+0{\color{gray}$\pm$1.9e-1} \\
  Sb      &       1.13e-6{\color{gray}$\pm$2.5e-7}   &  \best9.89e-7{\color{gray}$\pm$2.3e-7} &       1.02e-6{\color{gray}$\pm$2.7e-7} &       1.02e-6{\color{gray}$\pm$2.5e-7} &       1.09e-6{\color{gray}$\pm$2.7e-7} \\
  Sn      &       1.51e-8{\color{gray}$\pm$1.5e-9}   &       1.50e-8{\color{gray}$\pm$1.6e-9} &       1.20e-8{\color{gray}$\pm$1.4e-9} &       1.26e-8{\color{gray}$\pm$2.0e-9} &  \best9.41e-9{\color{gray}$\pm$1.1e-9} \\
  Sr      &       4.79e-4{\color{gray}$\pm$2.1e-4}   &       5.45e-4{\color{gray}$\pm$1.5e-4} &  \best3.84e-4{\color{gray}$\pm$1.2e-4} &       5.93e-4{\color{gray}$\pm$2.6e-4} &       5.81e-4{\color{gray}$\pm$2.4e-4} \\
  Zr      &       9.42e-4{\color{gray}$\pm$4.2e-4}   &       1.35e-3{\color{gray}$\pm$4.8e-4} &       9.14e-4{\color{gray}$\pm$2.5e-4} &       1.81e-3{\color{gray}$\pm$6.1e-4} &  \best8.64e-4{\color{gray}$\pm$2.2e-4} \\
\bottomrule
\end{tabular}

    }
      \caption{
        MSE (mean squared error) and standard error reported across cross-validation test sets.
        Best scores for each element across all models are highlighted.
      }
    \label{tbl:eval}
\end{table}

\begin{figure*}[!h]
 \begin{center}
  \includegraphics[width=1\linewidth]{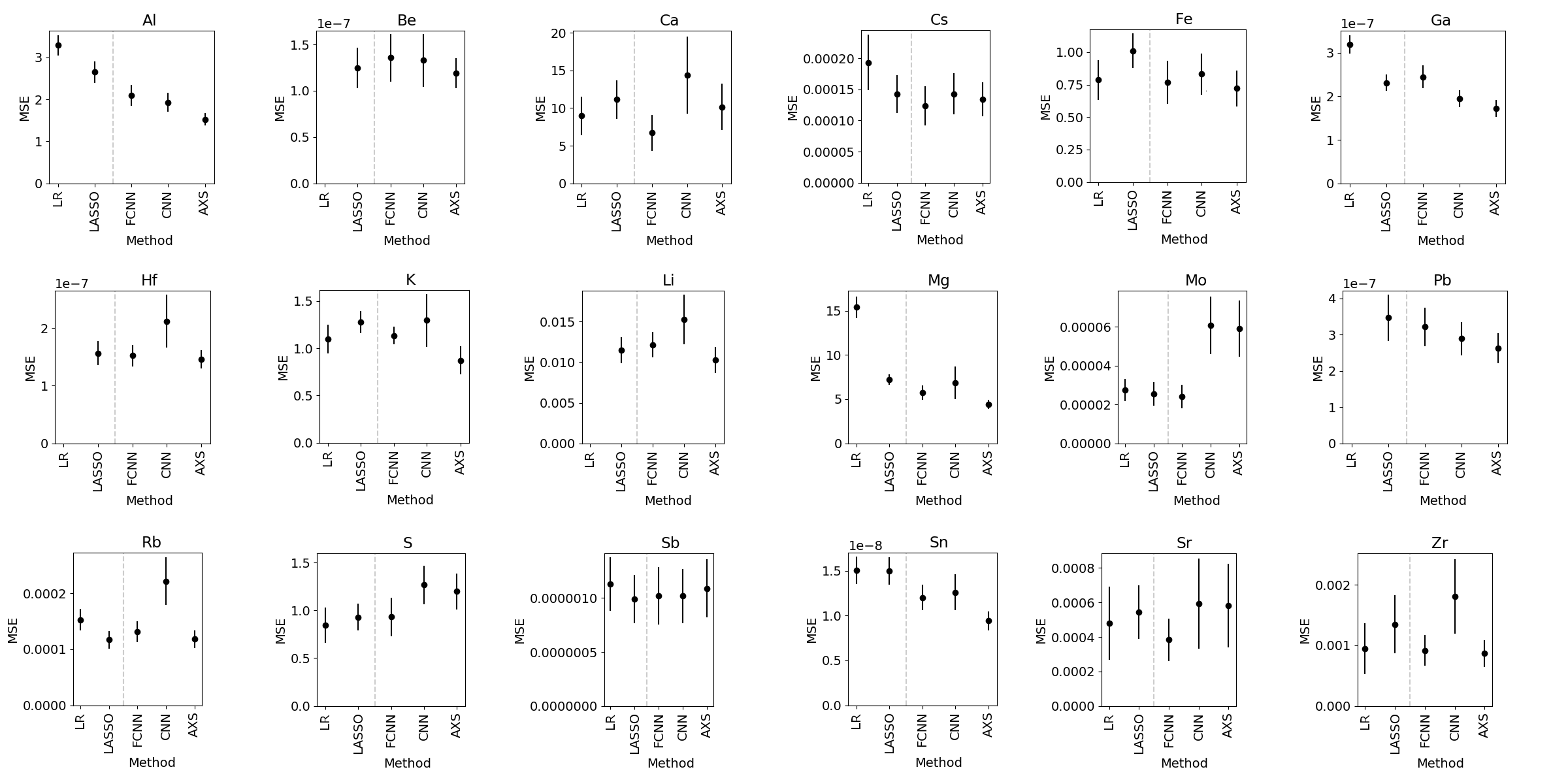}
  \caption{MSE and standard error (error bars) for each model and each element.}
  \label{fig:errorbars}
 \end{center}
\end{figure*}

\end{document}